\pdfoutput=1
\documentclass[12pt]{article}
\usepackage{amsmath}
\usepackage{amssymb}
\usepackage{amsthm}
\usepackage{amscd}
\usepackage{amsfonts}
\usepackage{graphicx}%
\usepackage{fancyhdr}

\theoremstyle{plain} \numberwithin{equation}{section}

\usepackage[margin=1in]{geometry}

\title{Approximate Query Matching for Image Retrieval}
\author{Abhijit Suprem \\ School of Computer Science, Georgia Tech}

\begin{document}

\maketitle

\section{Introduction}

\begin{figure}[h]

  \centering
    \includegraphics[width=.95\textwidth]{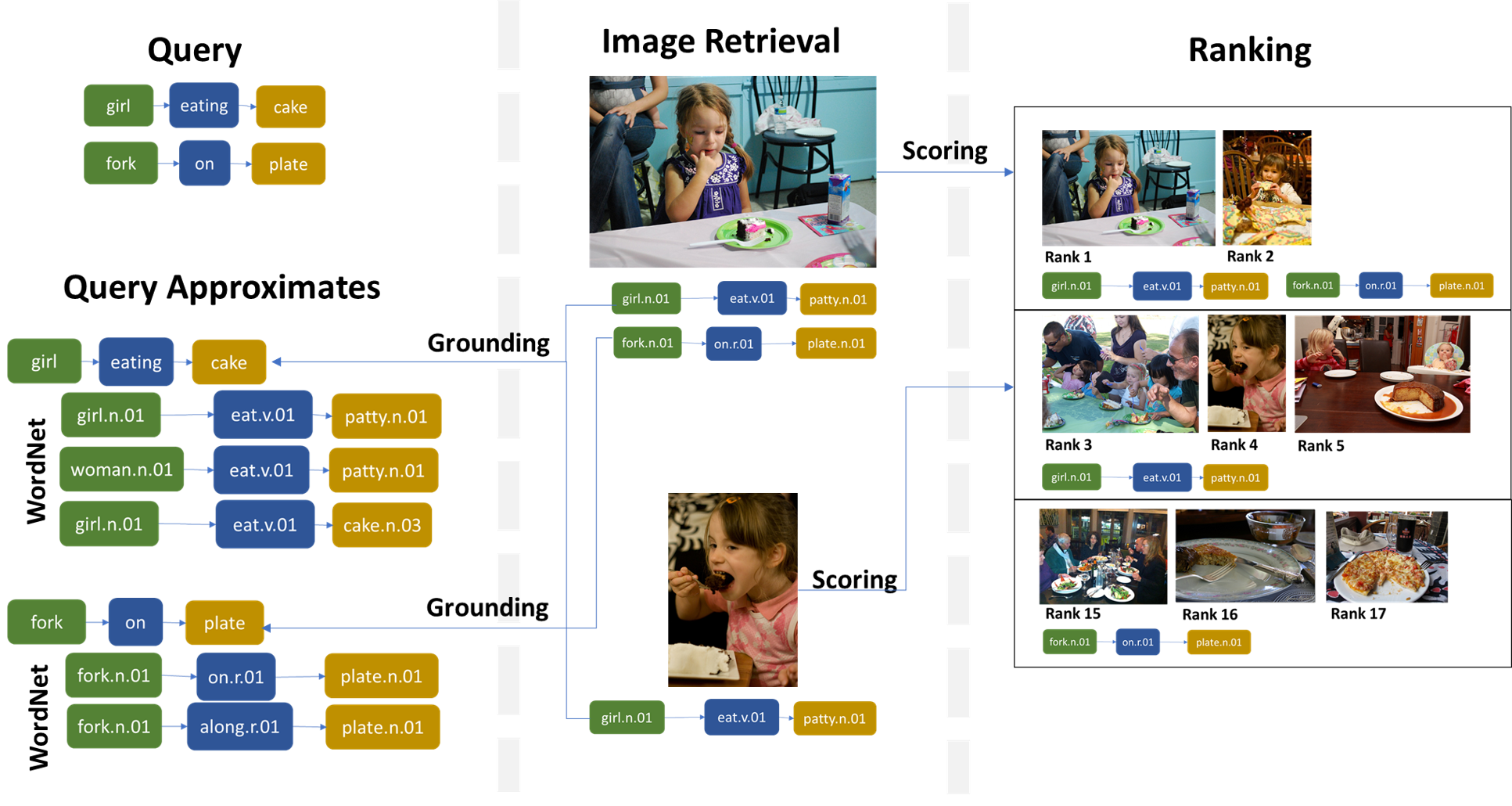}
    \caption{The user provides a query with two independent requests: \texttt{girl - eating - cake} and \texttt{fork - on - plate}. We generate queries that approximately match the given query  using WordNet synsets (\texttt{girl.n.01 - eat.v.01 - patty.n.01}, \texttt{fork.n.01 - along.r.01 - plate.n.01}). We then retrieve images that contain these subgraphs in their scene graphs, and ground the subgraphs to our top-level natural language queries. We then map the retrieved synsets and the query nodes to the same word embedding space, and measure similarities to determine holistic image similarity. the image similarity scores are then used to rank the images.}
\label{fig1}
\end{figure}

Given the explosion of image-based content shared digitally, there has been similar research focus on developing object recognition, image recognition and variants, and image segmentation tools and systems for understanding, identifying, storing, and querying images. The major research thrusts include automatic image captioning, image annotation (including relationship detection), and content-based image retrieval. Automatic image captioning attempts to generate natural language captions for image images that try to represent the holistic meaning of the image. As the saying goes, `A picture is worth a thousand words', and finding the right words is a non-trivial task. Image annotation focuses on object detection combined with relationship detection and grounding to identify canonical relationships within an image. Following \cite{kzg16}, these are of the form \texttt{<subject - predicate - object>}. Finally, content-based image retrieval is a wide net; we can consider some recent work from \cite{jks15}, who focus on using automatic image annotation via scene graph grounding to retrieve images similar to a query scene graph.

\begin{figure}[t]

  \centering
    \includegraphics[width=0.95\textwidth]{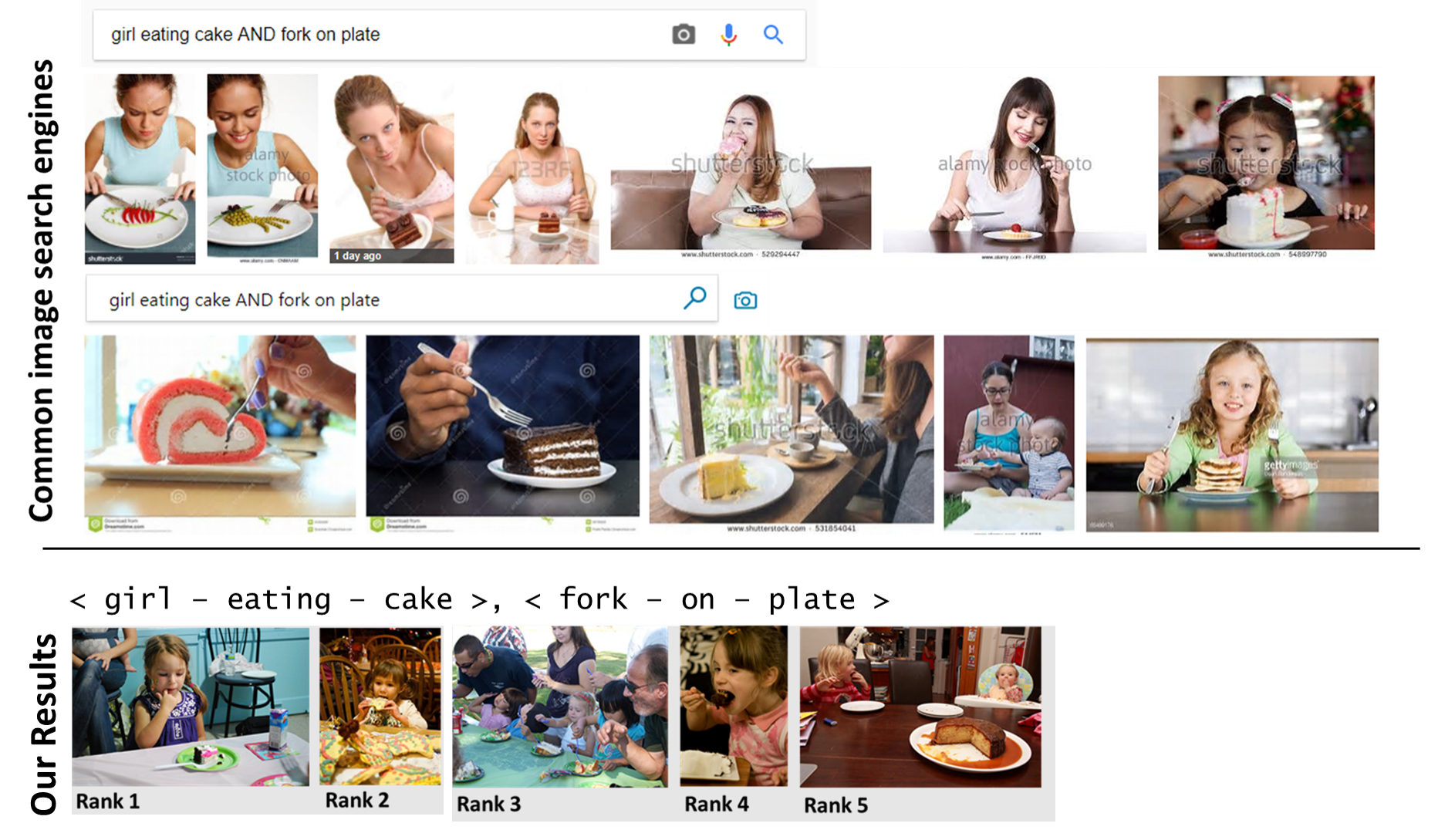}
    \caption{We show the results from common search engines for the same query. Our results are also presented. We note int he first set of results that the first two images do not include cake - in fact, the woman is eating peas and peppers. A woman eating cake appears third. In the second set of results, a profile view of a woman eating cake appears third in the results. Our results show the first two results with a woman (approximated as girl, as our far smaller image database did not contain any images that exactly matched the query) eating cake, with a fork on the plate.}\label{fig2}
\end{figure}

\subsection{Motivation}
The motivation for this work comes from the need for holistic image retrieval systems. As \cite{lzlm07} notes, text-based queries for image retrieval encode high levels of reasoning and abstraction about an image that is difficult to represent textually. In addition, text-based queries necessarily represent a semantic gap --- a disconnect between human semantics and visual features. \cite{jks15} demonstrates some drawbacks of current text-based image retrieval systems, namely the inability of such systems to abstract queries beyond index-based searches by recognizing the relationships between terms. We show this in Figure~\ref{fig2}, where common image search utilities return results that may not match the given query. We note in the first search engine's results that the second through fifth results feature a woman eating a cake, while in the second, none of the first three feature a girl eating a cake. Our results, however, return images that contain both features of the query, where possible, at least one (image 4 in our results). 

An overview of the process is shown in Figure~\ref{fig1}. Given the query , we extract the canonical forms - what we deem to be the smallest query unit. In the Visual Genome database, this is considered to be a subgraph of the form \texttt{<subject - predicate - object>}, where each of the \textit{subject}, \textit{predicate}, and \textit{object} are nodes with edges between them labeled with the type of relation (subject-to-predicate or predicate-to-object). We the generate approximates of the canonical forms to broaden our search. In this case, approximates of \texttt{<girl - eating - cake>} include \texttt{<girl.n.01 - eat.v.01 - patty.n.01>} and \texttt{<girl.n.01 - eat.v.01 - cake.n.03>}. Note that these approximates are represented as WordNet synsets as they are sued to search the Visual Genome database, whose scene graphs also represent nodes as WordNet synsets. We retrieve images and ground them to the query --- a process during which we identify which portions of the query graph the image satisfies and to what degree --  using phrase similarities derived from high dimensional word embeddings (we use LexVec \cite{siv16} for its state-of-the-art performance on Word Similarity measurements), Finally, we score each image on its grounding and rank the results.

Our work partially bridges the semantic gap by leveraging the idea of a scene graph as presented in the Visual Genome dataset \cite{kzg16}. A key contribution of \cite{kzg16} is the formalization of a scene graph --- each image in the Visual Genome Dataset contains a human-generated graph of nodes denoting subjects, objects, and predicates, with the edges denoting the relations between them. The graph annotates the relational content in the image. The dataset also contains region-level and image-level natural language captions, along with WordNet synsets for objects and relationships. There is a preponderance of research surrounding the scene-graph formalization --- Relationship Detection \cite{lkbff16}, Dense Caption Generation \cite{jkff16}, and Scene Graph Generation \cite{xzcff17}. We propose leveraging existing capabilities showcased in the above works for an efficient and scalable image retrieval pipeline.
Our contributions include:

\paragraph{Query approximation and ranking} We present a model for performing approximate search on a scene graph database. Given a query of the form \texttt{<subject - predicate - object>} (e.g. \texttt{<subject girl - eating - cake>}, we generate approximate queries searching for terms similar to the query terms and evaluating phrases for their similarities to the query. Using this, we can generate, for the given example, query approximates that include \texttt{woman - eating - cake}, \texttt{person - eating - cake}, \texttt{girl - eating - patty}, and \texttt{woman - eating - patty}. In addition, we present an algorithm for grounding and evaluating retrieved scene graphs to the query graph for holistic image ranking.

\paragraph{Aggregate graph representations} We present several graph database representations that are used in generating plausible query approximates by reducing the search space to the current semantic context. Specifically, we show how to create three aggregate graph representations that each index a different query node type: (i) a \textit{Subject Aggregate Graph} (SAG) for obtaining subject approximates, (ii) an \textit{Object Aggregate Graph} (OAG) for obtaining object approximates, and (iii) a \textit{Predicate Aggregate Graph} (PAG) for obtaining predicate approximates, all in the current semantic context.

\subsection{The Image Retrieval Pipeline}

\begin{figure}
	
  \centering
    \includegraphics[width=0.95\textwidth]{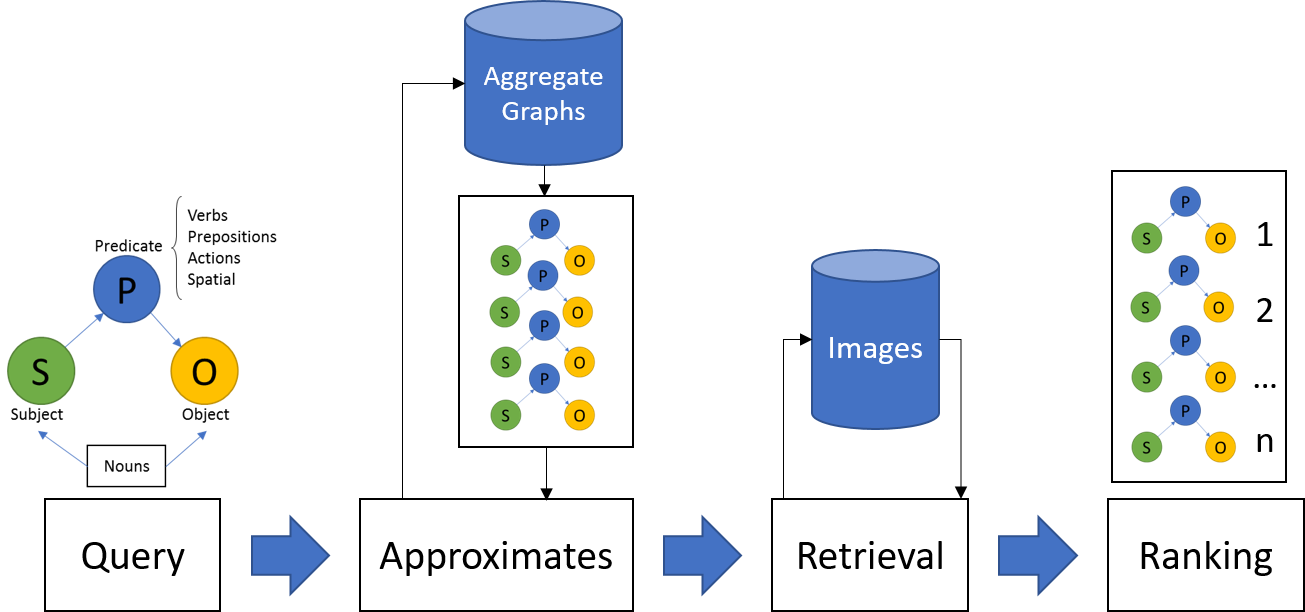}
    \caption{The user provides a natural language graph query. We convert this to the canonical form (the basic query unit of the form \texttt{<subject - predicate - object>}) for each top level query and generate the approximates. Plausible subject, object, and predicate approximates are generated using the aggregate graphs we have devised. These aggregate graphs --- the Subject Aggregate Graph, the Object Aggregate Graph, and the Predicate Aggregate Graph --- allow us to reduce the search space of subjects, objects, and predicates, respectively. Using the node approximates, we generate query approximates and retrieve the set of candidate images using an inverted index of the canonical queries from the Visual Genome dataset. We then ground the images to determine and rank how well the image represents our natural language query.}\label{fig3}
\end{figure}

Figures~\ref{fig1} and ~\ref{fig3} summarize our approach. We now focus on the pipeline in Figure~\ref{fig3}. Given a natural language graph query (e.g. a small scene graph as considered by \cite{jks15}), we reduce the query to its canonical forms (Section~\ref{canonical}); we define a canonical form as the basic unit of query consisting of a subject, predicate, and object of the form \texttt{<subject - predicate - object>}. We then generate approximate queries using WordNet synsets; we retrieve the set of candidate images that contain these approximates and ground the images' scene graphs to the query. We then use word embeddings and project each canonical form to the vector space and score each image's scene graph to the query graph, penalizing images with missing or inexact results.

\begin{figure}

  \centering
    \includegraphics[width=0.9\textwidth]{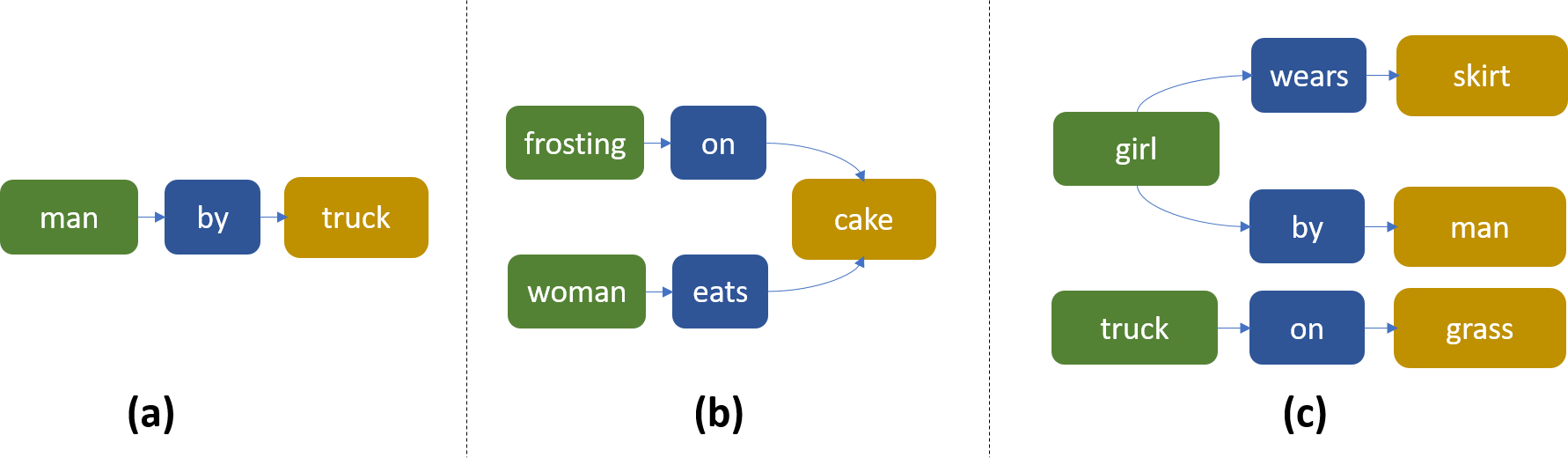}
    \caption{(a) A simple query. This is also the canonical form described in Section~\ref{canonical}. (b) A more complex query. The same \textbf{noun} (\textit{cake}) is the \texttt{object} of two separate \texttt{subjects} (\textit{woman} and \textit{frosting}), each connected by a different \texttt{predicate}. (c) Two independent top-level queries for the same image, i.e. they do not share any nodes between them. This consists of a `simple' top-level query and a `complex' top-level query, as per the informal terms adopted in (a) and (b).}\label{fig4}
\end{figure}

\begin{figure}[t]

  \centering
    \includegraphics[width=0.9\textwidth]{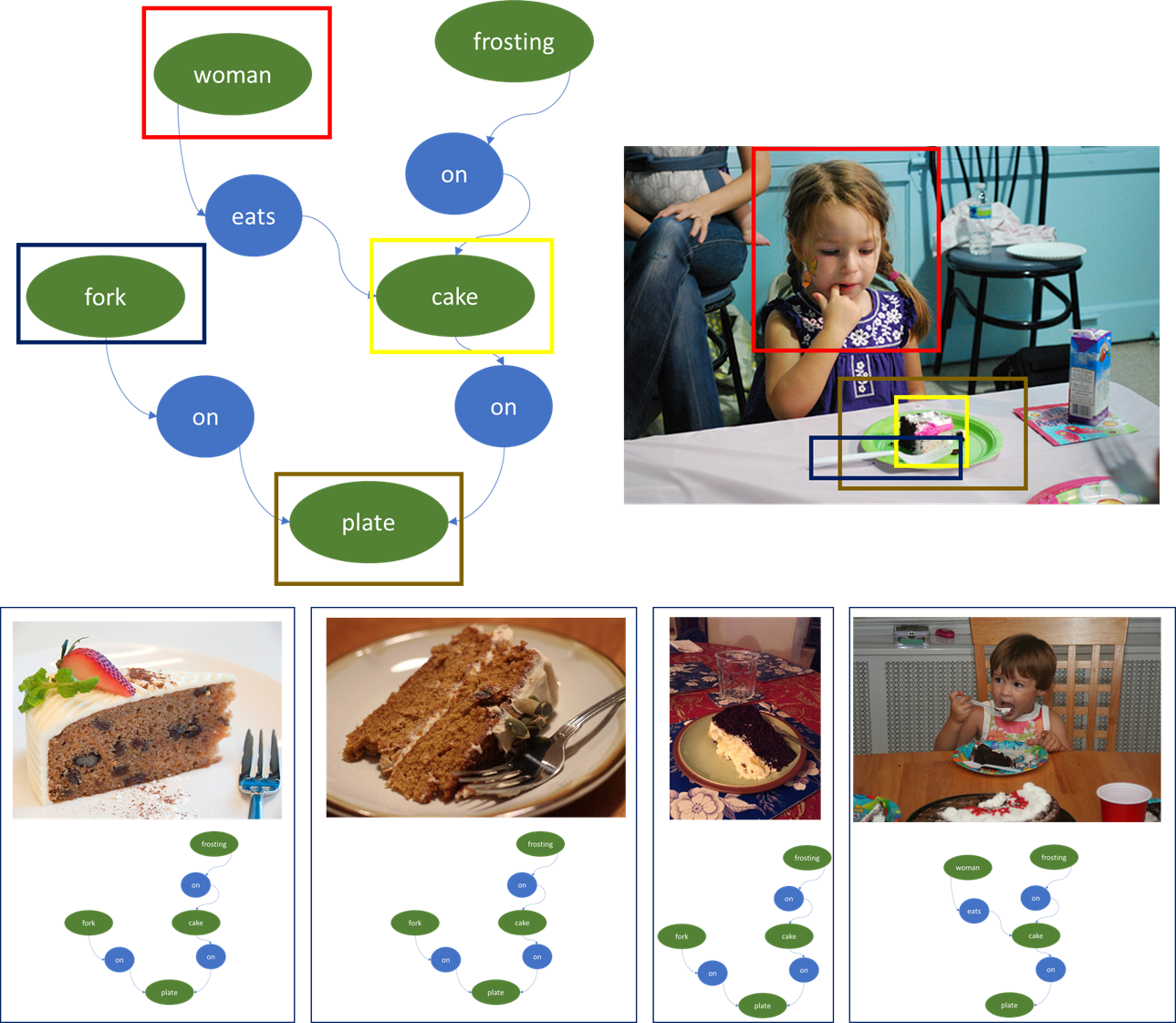}
    \caption{We show our system matching a complex query and returning images that most closely match all facets of the query.}\label{fig5}
\end{figure}

Three sample queries of varying complexity are presented in Figure~\ref{fig4}. Our approach handles both simple and complex queries. Our grounding ensures images that most closely match the entire query are ranked higher than images that offer a partial match. In addition, we bridge the `semantic gap' by searching for approximates. This allows us to return images that closely match the provided query in situations where an exact match may not be possible; an example is shown in Figure~\ref{fig5}: there is no image that exactly matches the query; however, our approach approximates woman into girl, and finds images that contain most of the query: a girl eating cake, with the cake on a plate, and a fork on the plate as well. While the highest ranked image in the figure does contain cake with frosting, this is a happy coincidence; the scene graph itself does not contain the annotation, so we say the image mostly matches the query. We also show the second through fifth ranked images and their matches scene subgraphs.

\section{Related Work}

\paragraph{Content-based image retrieval}
Content-based image retrieval involves retrieving and ranking images given either a text or image query. The former can be used to retrieve images that incorporate some of the query requirements, while the latter can be used to retrieve images  relevantly similar to the query --- here evaluation is context-specific, e.g. whether images have similar objects, similar colors, similar poses, or whether the images are exactly the same. Some implementations include Google Reverse Image Search and TinEye. Our focus is on the former approach, as we feel text-based queries allow users more freedom in specifying images to retrieve. It is not the case that a user always has a sample of the image she would like, or can sketch a faithful representation in, e.g., sketch-based image retrieval. \cite{jks15}'s key contribution is as follows: the authors develop a framework for generating accurate groundings of a query scene graph (either complete or partial), and use grounding likelihood to rank images for retrieval and display. 

\paragraph{Semantic Similarity}
An integral aspect of our image retrieval pipeline is approximate query generation. We work with canonical relationship phrases of the form \texttt{<subject - predicate - object>} to generate query approximates. As the Visual Genome dataset maps each object to available WordNet synsets, we incorporate WordNet based semantic similarity measures. \cite{pp06} proposes a domain-specific corpus-based training method to identify correct word sens and derives more accurate cosine-similarity measures between source and target words. \cite{krfa17} shows a simple baseline for WordNet synset similarity using vector embedding cosine similarity averages. \cite{alm17} shows a sentence-based similarity measure that uses a TF-IDF analogue to compute similarities between a source word and its synset lemmas.

\section{Graph Databases}
As noted, \cite{kzg16} formalizes the scene graph --- an image representation using human annotations on bounded regions that grounds objects, relationships, and attributes. Each object (usually a noun) is considered a node with a directed edge towards a relation (a predicate with a part-of-speech tag of verb, preposition, or action), The predicate may or may not have a directed edge towards another object instance. In addition, objects also have attributes (usually adjectives, but may also include actions). We will henceforth consider each \textit{object} node as a \textbf{noun} with two forms: \texttt{subject} or \texttt{object}. Note that \textit{object} and \texttt{object} operate under separate domains; however notation confusion suggests these terms as an appropriate choice. A canonical form is a set of three nodes and two edge of the form \texttt{<subject $\rightarrow$ predicate $\rightarrow$ object>}: this triplet indicates a base query that we use as the smallest query unit (i.e. the canonical form). 

\subsection{Full Scene Graphs}
We generate the full scene graph for each image using the \texttt{scene\_graphs} from the Visual Genome dataset. The full scene graph is stored in a Neo4J database. Each full scene graph consists of at least one subgraph with at least one canonical triplet of the form \texttt{<subject - predicate - object>}. We note that there may be multiple independent subgraphs corresponding to different regions in the image.

\subsection{Object Aggregate Graph}
\label{oag}
Our query approximation requires us to generate the candidate subjects, objects, and predicates for each top-level query. We develop a novel aggregate characterization of the aggregate graph that is introduced in \cite{kzg16} --- we maintain a unique index of \texttt{<subject $\rightarrow$ predicate>} pairs, and for each pair, we maintain a unique list of \texttt{objects} that are associated with that \texttt{subject$\rightarrow$predicate} pair. \texttt{Objects} are not unique across each subgraph in the aggregate graph: \texttt{apple.n.01} may appear in multiple \texttt{<subject $\rightarrow$ predicate>} subgraphs; however within a subgraph, each WordNet synset occurs once.

\begin{figure}[h]

  \centering
    \includegraphics[width=0.6\textwidth]{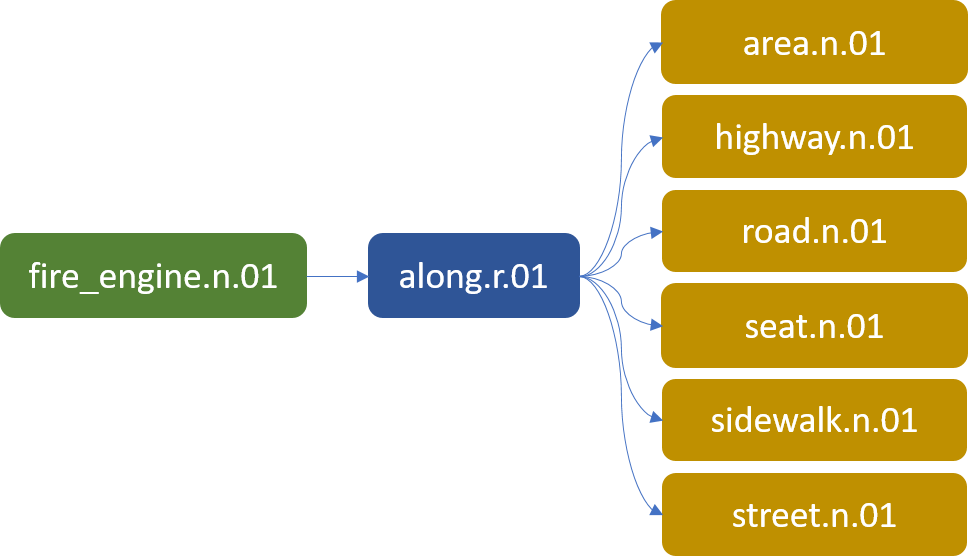}
    \caption{This subgraph from the Object Aggregate Graph shows all possible objects that a Fire Engine (\texttt{fire\_engine.n.01}) can be next to (\texttt{along.r.01}). These include streets (\texttt{street.n.01}), sidewalks (\texttt{sidewalk.n.01}), and highways (\texttt{highway.n.01}). Given a set of approximate synsets for \textit{street}, as well as the \texttt{fire\_engine.n.01$\rightarrow$along.r.01} pair, we can identify the appropriate set of plausible objects using the process from Section~\ref{oag}.}\label{fig7}
\end{figure}

With the \textbf{Object Aggregate Graph}, we can, given a \texttt{subject$\rightarrow$predicate} pair as well as several candidate \texttt{objects}, reduce the set of candidate \texttt{objects} to plausible candidates for that \texttt{subject$\rightarrow$predicate}. Given candidate objects $C_O$ and child objects $K_O$ in the \textbf{Object Aggregate Graph}, we return $C_O\cup K_O$. Figure~\ref{fig7} shows a subgraph within our \textbf{Object Aggregate Graph}

\subsection{Subject Aggregate Graph}
\begin{figure}[t]

  \centering
    \includegraphics[width=0.6\textwidth]{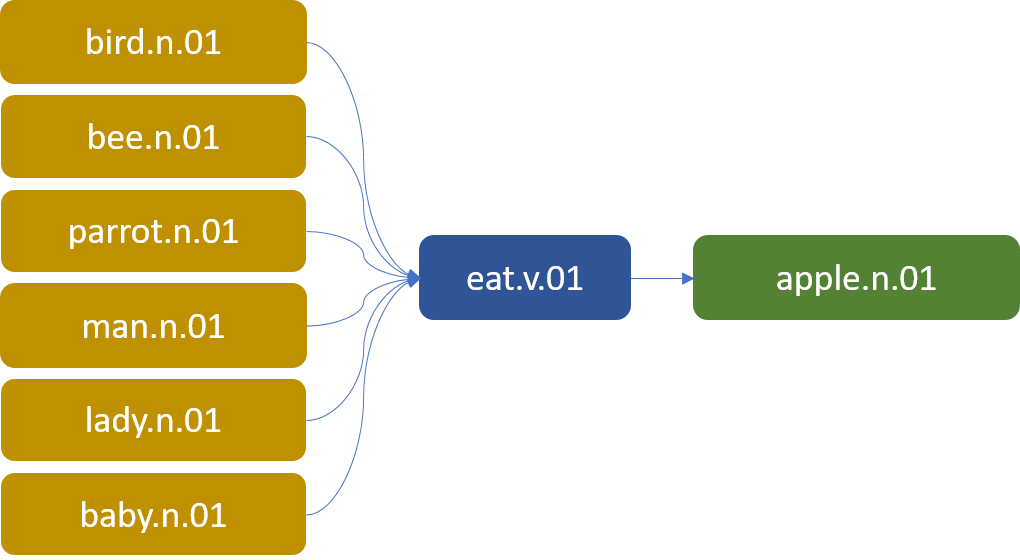}
    \caption{This subgraph from the  Subject Aggregate Graph shows all possible objects that can eat (\texttt{eat.v.01}) an apple (\texttt{apple.n.01}). These include woman (\texttt{woman.n.01}), child (\texttt{child.n.01}), and girl (\texttt{girl.n.01}).}\label{fig8}
\end{figure}

To generate candidate subjects, we devise a \textbf{Subject Aggregate Graph}: we maintain a set of unique \texttt{predicate$\rightarrow$object} pairs, and for each pair, we keep a unique set of \texttt{subjects} that are associated with that pair. So, given a \texttt{predicate$\rightarrow$object} pair, and candidate subjects $C_S$, we reduce it to the plausible set of subjects by returning $C_S\cup K_S$ (where $K_S$ are the parent objects in each subgraph in the Subject Aggregate Graph). We show this in Figure~\ref{fig8}.

\subsection{Predicate Aggregate Graph}
\begin{figure}

  \centering
    \includegraphics[width=0.6\textwidth]{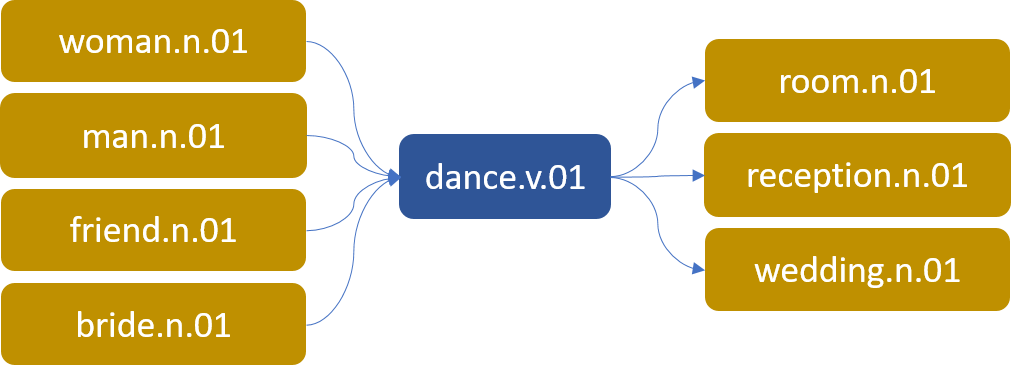}
    \caption{This partial predicate aggregate graph shows all possible nodes that are subjects and objects for the predicate \textit{dance} (\texttt{dance.v.01}). This allows us to determine plausible predicates given given a subject-object pair.}\label{fig9}
\end{figure}

We adapt \cite{kzg16}'s aggregate graph here; however, we include all \textbf{nouns} and \textbf{predicates} from the scene graphs, instead of the top-\textit{k} \textbf{nouns} and \textbf{predicates}. Given a set of candidate \texttt{subjects} and candidate \texttt{objects}, we can obtain the set of plausible predicates, and return the union of the candidate and plausible predicates. We show a subgraph in Figure~\ref{fig9}.

\section{Query Matching and Image Retrieval}
\subsection{Canonical Form}
\label{canonical}
The canonical form of a query is a triplet (Figure 4a) of the form \texttt{<subject $\rightarrow$ predicate $\rightarrow$ object>}. Given a graph query, we extract canonical forms from the query and operate on each independently, within scope of the query subgraph. Once again we refer to Figure 4c --- this query consists of two independent subgraphs: one about a \texttt{girl by a man, with the girl wearing a skirt} and one about a \texttt{truck on grass}. We split the first query into \texttt{<girl  $\rightarrow$ wears $\rightarrow$ skirt>} and \texttt{<girl$\rightarrow$ by $\rightarrow$man>}. Similarly, we extract from \texttt{truck on grass} the triplet \texttt{<truck$\rightarrow$on$\rightarrow$grass>}.

We perform this triplet extraction under an independence assumption --- that we can extract individual triplets, the combine them later to obtain the final rankings. This allows us to operate independently on each triplet and its set of query approximates.

\subsection{Approximate Generation and Retrieval}
Given a triplet, we obtain the WordNet approximates for the subject, predicate, and object. There are three levels of approximates --- obtaining the sister synsets, the child synsets (hyponymy), and the parent synsets (hypernymy). The sister synsets are obtained by performing a lookup in the WordNet database of out natural language node label (i.e. `girl' or `skirt' or `wears'). there are four scopes available for synset lookup: we can take (i) just the sister synsets, (ii) the sister and child synsets, (iii) the sister and parent synsets, (iv) or all three hierarchies: sister, child, and parent synsets. We limit our choice for candidate subjects and objects to the sister synsets and for candidate predicates to sister and child synsets. This is to speed up computation time on our local machines; on parallelized clusters, such a limitation is not necessary and we can use the complete closure of a synset: sisters, children, and parents.

Given the set of approximates for the subject and object, we obtain plausible predicates from the aggregate graph. The predicates are ranked to the provided predicate sister synsets: we measure the Wu \& Palmer (WUP) Similarity between the plausible predicates and set of sister predicates, and taker the average similarity score. Here, we take the top 2/3 predicates as our set of predicate approximates.  We choose WUP similarity as it weights synset edges by distance in the hierarchy, i.e. semantically similar predicates are ranked higher than synonymy predicates without semantic relation. This allows us to narrow results to more appropriate relations by context.

\subsection{Synset Embeddings}
For ranking, we prefer to use a Euclidean metric for measuring triplet distance between the query and approximates. However, the WordNet hierarchy does not operate on such a metric. We use a model similar to \cite{krfa17} --- we determine the embeddings for each synset in our database following the baseline method for \cite{krfa17}, but instead of the sum, we take the average of the embeddings sum to obtain the synset centroid in the embedding space with 
\begin{equation}
\label{eq1}
v_S=\frac{1}{|L_S|}\sum_{l\in L_s}v_l
\end{equation}

where $v_S$ is the embeddings vector for a synset $S$, $L_S$ is the set of lemmas for the synset $S$, and $V_l$ is the embedding for each lemma $l\in L$. We use LexVec \cite{siv16} embeddings as the lookup table for $v_l$ as LexVec has shown state-of-the-art performance in word similarity and analogy. We deal with compound words by averaging the embeddings for the compound word itself and the embeddings sum of its component words. If the compound word does not exist, we take only the embeddings sum of component words: given a compound word $w = w_1, w_2,\cdots w_n$ (i.e. for $w=$\texttt{christmas tree}, $w_1=$ \texttt{christmas} and $w_2=$ \texttt{tree}), we take $v'_l=\frac{1}{2}\big( v_l+\sum_{l={l_1,l_2,\cdots,l_n}}v_{l_i}\big)$.

\subsection{Image Ranking}

\subsubsection{Approximate-triplet image retrieval}
\begin{figure}

  \centering
    \includegraphics[width=0.95\textwidth]{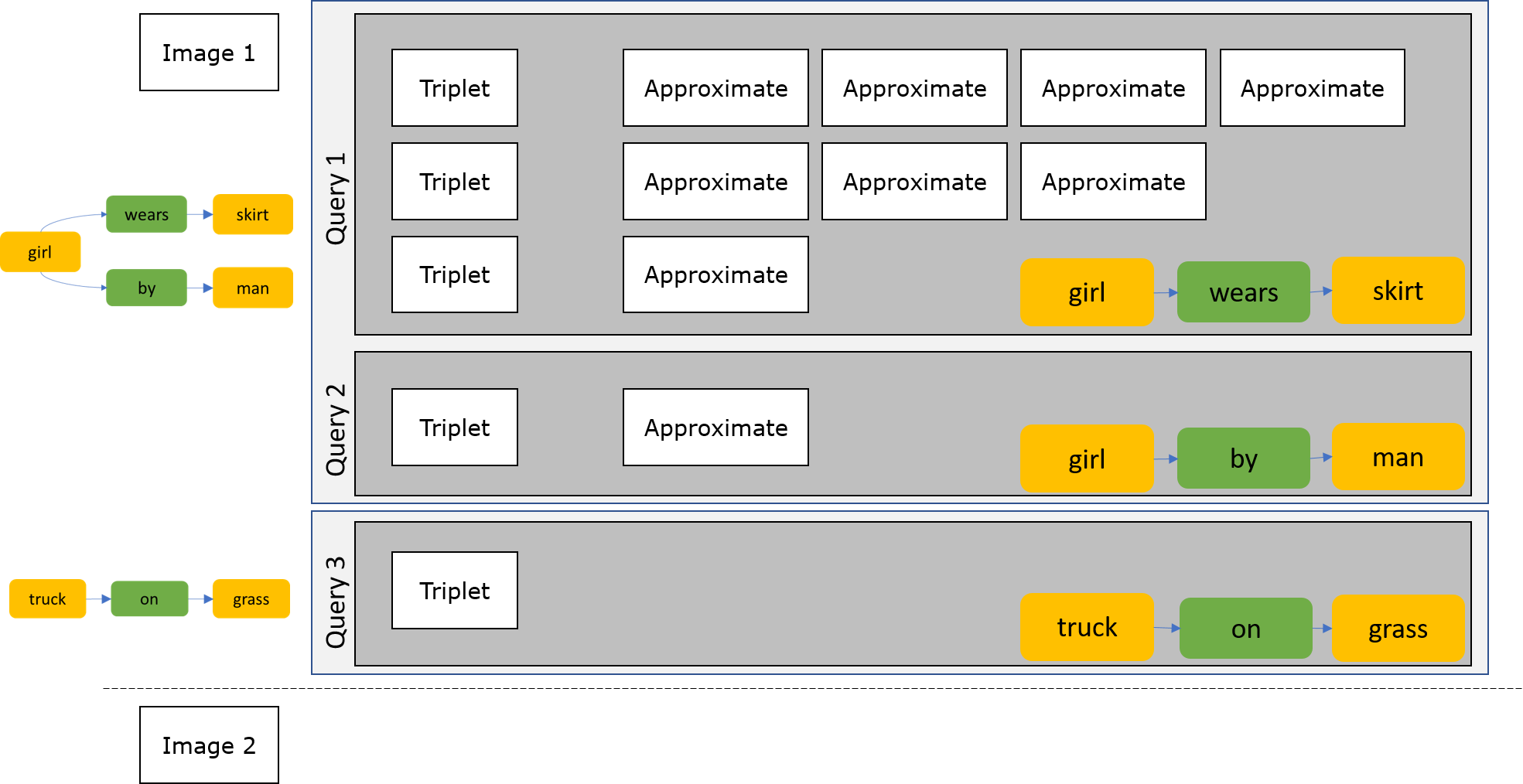}
    \caption{We store all approximates for each triplet for each image. It is possible that images may not contain a relation, in which case the index will not contain any approximate for a given query (\texttt{truck on grass}).}\label{fig10}
\end{figure}

After triplet retrieval, we have, for each canonical triplet $T_i$, a set of approximates $A^{(T_i)}=a_1, a_2, \cdots, a_k$. Each of these approximate is a triplet that is semantically similar to the query triplet. For each approximate $a_i$, we have a set of images $I_1, I_2, \cdots , I_n$ that contain the approximate triplet within them. We generate an inverted index of images for for each image, we collect all approximates $a_k$ for each triplet $T_i$ (Figure~\ref{fig10}). This allows us to work on an image-by-image basis by scoring each image on the basis of its holistic similarity to the query triplet.

\subsubsection{Inverted Index}
Each approximate is a triplet of synsets, and we obtain the synset embeddings using the method in Equation~\ref{eq1}. We also obtain the embeddings for the canonical representations from the embeddings lookup (LexVec, in this case). At this point, we note that each approximate triplet is  independent of every other approximate triplet. We further note that each approximate $a_i\in I_n$, there may be multiple triplets that match the approximate. Consider an approximate triplet of the form \texttt{man on grass}: an image may contain multiple instances of this approximate, and each is stored in the inverted index. Each of these approximate primitives $ap_j$ contains a grounding of its nodes to nodes in the parent query approximate $a_k$, which itself contains a grounding to its parent triplet $T_i$. As such, for each $ap_{i\in [1,j]}$, we know what it grounds to in $T_i$. We further note that a single image triplet may appear in multiple primitives $ap_j$ as the parent queries could have similar triplets in multiple subgraphs.

\subsubsection{Subgraph scoring}
We first `collapse' the obtained triplets to their subgraphs $s_a$ --- we generate the image subgraphs that contain all approximate primitives within a  single top level query triplet $T_i = s_1, s_2, \cdots, s_a$. As such, we reduce all redundant copies of each unique node within the inverted index and connect the independent triplets wherever they share subjects or predicates. This is necessary for the query matching we will perform to rank each image approximate to the provided query. We want a notion of subgraph isomorphism to match out collapsed primitives to our top level query $T_i$. From Figure~\ref{fig10}, we note that there  may be some $T_i$ that do not contain approximates. For each $s_a$, we obtain the cosine similarity between its synset embedding and its ground embedding, which is the embedding of the node in the top level query $T_i$. The cosine similarity is in $[0,1]$, where 1 indicates the highest similarity. For each node, we instead store $n_{score}=1-\mathtt{cosine\_sim}(node,query)$. For each missing nodes in $s_a$, we add a null node with a distance of 1, representing a missing node.

\subsubsection{Subgraph ranking}
With this representation, we can now formulate this as a minimization problem: we wish to select the subgraph with the smallest score. Since each subgraph $s_a\in T_i$ may contain a combination of approximates, it is intractable to calculate the global minimum selection of subgraphs. We instead model this as an analogue of vertex cover. By construction, each subgraph contains a unique set of approximates for the top level query. As such, larger subgraphs are inherently more `isomorphic' to each $T_i$ simple because they contain more grounded nodes. 

We then sort our subgraphs by size, and within each set of subgraphs with the same size, we pick the subgraph $s_m$ with the smallest score. This subgraph $s_m$ has a subset of the top level approximates in $T_i$. We remove these approximates from consideration, and again pick, from the remaining, the largest subgraphs. We repeat this until all top level queries are satisfied for each top level subgraph. Each of the selected subgraphs $S_m$'s scores $S_i=score(s_1, s_2,\cdots, s_m)$ are summed and normalized by the number of nodes to find the score on $I_n$ for query $T_i$. We repeat this for each query in $I_n$ to obtain scores $S={S_1,S_2,\cdots, S_n}$ for each query $T_i$.

\subsubsection{Image scoring and ranking}
We note that for each of the scores $S={S_1,S_2,\cdots, S_n}$  is in [0,1], with a lower score corresponding to a more similar match for each top level query. It is straightforward then to represent these scores as an $i$-dimensional score vector $[S_1,S_2,\cdots,S_i]$ and measure its Euclidean distance from the origin. This gives us a score for image $I_n$. We then sort the image score to obtain the image ranking under query, where the smallest score is the most relevant image.

\section{Conclusions and Future Work}
We have presented work on image retrieval using graph based approximate querying and ranking. Representative examples are shown in Figures~\ref{fig1} and ~\ref{fig5}, as well as a comparison in Figure~\ref{fig2}. We note from Figure~\ref{fig2} that our system returns relevant results at higher ranks than two leading Search Engines. We also note from Figure~ref{fig5} that even for more complex queries, out system can return results that closely match the provided query. Top ranked results match as much of the query as possible \textit{with holistic meaning} --- we reduce the `semantic gap' by considering relations between objects in the image in lieu of using a document-based that eschews a focus on image relational content. 

As such, the scene graph of an image is a key factor of our work. With regard to this, future work is two-fold:

\begin{itemize}
\item Accurate scene graph generation: There is some work on scene graph generation in \cite{xzcff17}. However, the authors note that the performance is subpar. Better state-of-the-art performance in automated scene graph generation from unannotated images would allow creation of image databases at scale. Our system can then be implemented on top of such a database for approximate image retrieval.
\item Query graph generation: We provide an informal  comparison of our results to current search engine results in Figure~\ref{fig2}. However, this is not a robust comparison as search engines accept natural language input while we provide input directly as graph queries. This is due to a major input limitation in the conversion of natural language to scene graphs. Future work would focus on graph query generation from natural language that more closely matches desired human queries, using, e.g. dependency parsing or similar methods.
\end{itemize}

\bibliographystyle{alpha}
\bibliography{main.bib}

\end{document}